\pdfoutput=1

\documentclass[11pt]{article}

\usepackage{ACL2023}

\usepackage{times}
\usepackage{latexsym}

\usepackage[T1]{fontenc}

\usepackage[utf8]{inputenc}

\usepackage{microtype}

\usepackage{inconsolata}

\usepackage{graphicx}
\usepackage{amsmath}
\usepackage{amssymb}
\usepackage{multirow}
\usepackage{todonotes}
\usepackage{xcolor}
\usepackage{booktabs}
\usepackage[T1]{fontenc}
\usepackage[utf8]{inputenc}
\usepackage{tabularx}
\usepackage{pifont}
\usepackage{enumitem}

\include{math}

\newcommand{\mypara}[1]{\noindent\textbf{#1}}
\newcommand\bund[1]{\textbf{\underline{#1}}}

\newcommand\und[1]{\underline{#1}}
\newcommand{\cmark}{\ding{51}}%
\newcommand{\xmark}{\ding{55}}%

\DeclareMathOperator*{\argmax}{arg\,max}

\definecolor{ForestGreen}{rgb}{0.56, 0.74, 0.56}
\definecolor{OliveGreen}{rgb}{0.1, 0.40, 0.25}

\usepackage{microtype}

\title{Context-Aware Transformer Pre-Training for Answer Sentence Selection}

\author{Luca Di Liello$^{1}$\thanks{\ \ Work done as an intern at Amazon Alexa AI}\ , Siddhant Garg$^{2}$, Alessandro Moschitti$^{2}$\\
$^{1}$University of Trento , $^{2}$Amazon Alexa AI\\
\texttt{luca.diliello@unitn.it} \\ \texttt{\{sidgarg,amosch\}@amazon.com} \\
}

\begin{document}
\maketitle

\begin{abstract}
\vspace{-.4em}
Answer Sentence Selection (AS2) is a core component for building an accurate Question Answering pipeline. AS2 models rank a set of candidate sentences based on how likely they answer a given question. The state of the art in AS2 exploits pre-trained transformers by transferring them on large annotated datasets, while using local contextual information around the candidate sentence. In this paper, we propose three pre-training objectives designed to mimic the downstream fine-tuning task of contextual AS2. This allows for specializing LMs when fine-tuning for contextual AS2. Our experiments on three public and two large-scale industrial datasets show that our pre-training approaches (applied to RoBERTa and ELECTRA) can improve baseline contextual AS2 accuracy by up to 8\% on some datasets.
\end{abstract}

\section{Introduction}
\vspace{-.2em}
\label{sec:intro}

Answer Sentence Selection (AS2) is a fundamental task in QA, which consists of re-ranking a set of answer sentence candidates according to how correctly they answer a given question. From a practical standpoint, AS2-based QA systems can operate under much lower latency constraints than corresponding Machine Reading (MR) based QA systems. Nowadays, latency is of particular importance because sources of information such as Knowledge Bases or Web Indexes may contain million or billion of documents. In AS2, latency can be minimized because systems process several sentences/documents in parallel, while MR systems parse the entire document/passage in a sliding window fashion before finding the answer span~\cite{garg-moschitti-2021-will,gabburo-etal-2022-knowledge}.

Modern AS2 systems~\cite{garg2019tanda,laskar-etal-2020-contextualized} use transformers to cross-encode question and answer candidates together. Recently, \citet{Lauriola2021} proved that performing answer ranking using only the candidate sentence is sub-optimal, for e.g., the answer sentence may contain unresolved coreference with entities, or the sentence may lack specific context for answering the question. Several works~\cite{shalini2016clstm,8226822,han2021modeling} have explored performing AS2 using context around answer candidates (for example, adjacent sentences) towards improving performance. Local contextual information, i.e., the previous and next sentences of the answer candidates, can help coreference disambiguation, and provide additional knowledge to the model. This helps to rank the best answer at the top, with minimal increase in compute requirements and latency.

Previous research works~\cite{Lauriola2021,han2021modeling} have directly used existing pre-trained transformer encoders for contextual AS2, by fine-tuning them on an input comprising of multiple sentences with different roles, i.e., the question, answer candidate, and context (previous and following sentences around the candidate). This structured input creates practical challenges during fine-tuning, as standard pre-training approaches do not align well with the downstream contextual AS2 task, e.g., the language model does not know the role of each of these multiple sentences in the input. In other words, the extended sentence-level embeddings have to be learnt directly during fine-tuning, causing under-performance empirically. This effect is amplified when the downstream data for fine-tuning is small, indicating models struggling to exploit the context.

In this paper, we tackle the aforementioned issues by designing three pre-training objectives that structurally align with the final contextual AS2 task, and can help improve the performance of language models when fine-tuned for AS2. Our pre-training objectives exploit information in the structure of paragraphs and documents to pre-train the context slots in the transformer text input.
We evaluate our strategies on two popular pre-trained transformers over five datasets. The results show that our approaches using structural pre-training can effectively adapt transformers to process contextualized input, improving accuracy by up to 8\% when compared to the baselines on some datasets.

\section{Related Work}
\label{sec:related_works}
\vspace{-.2em}

\paragraph{Answer Sentence Selection} TANDA~\cite{garg2019tanda} established the SOTA for AS2 using a large dataset (ASNQ) for transfer learning. Other approaches for AS2 include: separate encoders for question and answers~\cite{bonadiman-moschitti-2020-study}, and compare-aggregate and clustering to improve answer relevance ranking~\cite{compare1905}.

\paragraph{Contextual AS2} ~\citet{shalini2016clstm} use LSTMs for answers and topics, improving accuracy for next sentence selection. \citet{8226822} use GRUs to model answers and local context, improving performance on two AS2 datasets. \citet{Lauriola2021} propose a transformer encoder that uses 
context to better disambiguate between answer candidates. \citet{han2021modeling} use unsupervised similarity matching techniques to extract relevant context for answer candidates from documents. 


\paragraph{Pre-training Objectives} Pre-training sentence-level objectives such as NSP~\cite{devlin-etal-2019-bert} and SOP~\cite{lan2020albert} have been widely explored for transformers
to improve accuracy for downstream classification tasks. However, the majority of these objectives are agnostic of the final tasks. End task-aware pre-training has been studied for summarization~\cite{rothe-etal-2021-thorough}, dialogue ~\cite{li2020dialogueobjectives}, passage retrieval~\cite{condenser2021objectives}, MR~\cite{ram-etal-2021-shot} and multi-task learning~\cite{dery2021alternative}.
\citet{latent-2019}, \citet{pretraining-large-retrieval-2020} and \citet{end-to-end-2021} use the Inverse Cloze task to improve retrieval performance for bi-encoders, by exploiting paragraph structure via self-supervised objectives.
For AS2, recently \citet{multi-sentence2022} proposed paragraph-aware pre-training for joint classification of multiple candidates. \citet{diliello2022pretraining} propose a sentence-level pre-training paradigm for AS2 by exploiting document and paragraph structure. However, these works do not consider the structure of the downstream task (specifically contextual AS2). To the best of our knowledge, ours is the first work to study transformer pre-training strategies for AS2 augmented with context using cross-encoders.

\section{Contextual AS2}
\label{sec:methodology}
\vspace{-.2em}

\paragraph{AS2} Given a question $q$ and a set of answer candidates $S = \{s_1, \dots, s_n \}$, the goal is to find the best $s_k$ that answers $q$.
This is typically done by learning a binary classifier $\mathbf{C}$ of answer correctness by independently feeding the pairs $(q, s_i), i\in\{1,{\dots},n\}$ as input to $\mathbf{C}$, and making $\mathbf{C}$ predict whether $s_i$ correctly answers $q$ or not. At inference time, we find the best answer for $q$ by selecting the answer candidate $s_k$ which scores the highest probability of correctness $k = \argmax_i \mathbf{C}(q, s_i)$.

\paragraph{Contextual AS2} Contextual models for AS2 exploit additional context to improve the final accuracy. This has been shown to be effective \cite{Lauriola2021} in terms of overcoming coreference disambiguation and lack of enough information to rank the best answer at the top. Different from the above case, contextual AS2 models receive as input a tuple $(q, s_i, c_i)$ where $c_i$ is the additional context. $c_i$ is usually the sentences immediately before and after the answer candidate.

\section{Context-aware Pre-training Objectives}
\label{ssec:objectives}
\vspace{-.2em}

We design a transformer pre-training task that aligns well with fine-tuning contextual AS2 models, both \emph{structurally} and \emph{semantically}. We exploit the division of large corpora in documents and the subdivision of documents in paragraphs as a source of supervision. We provide triplets of text spans $(a, b, c)$ as model inputs when pre-training, which emulates the structure of $(q, s_i, c_i)$ for contextual AS2 models, where $a$, $b$ and $c$ play the analogous role of the question, the candidate sentence (that needs to be classified), and the context (which helps in predicting $(a, b)$ correctness), respectively. Formally, given a document $D$ from the pre-training corpus, the task is to infer if $a$ and $b$ are two sentences extracted from the same paragraph $P \in D$. Following \citet{diliello2022pretraining}, we term this task: ``Sentences in Same Paragraph (SSP)''.

\mypara{Intuition for SSP} Consider an example of a Wikipedia paragraph composed of three sentences:

\noindent $s_1$: Lovato was brought up in Dallas, Texas; she began playing the piano at age seven and guitar at ten, when she began dancing and acting classes.

\noindent $s_2$: In 2002, Lovato began her acting career on the children's television series Barney \& Friends, portraying the role of Angela.

\noindent $s_3$: She appeared on Prison Break in 2006 and on Just Jordan the following year.

Given a question of the type "What are the acting roles of $X$", a standard LM can easily reason to select answers of the type "$X$ acted/played in Y", by matching the subject argument of the question with the object argument of the answer, for the same predicate acting/playing. However, the same LM would have a harder time selecting answers of the type "$X$ appeared in $Y$" because this requires learning the relation between the entire predicate argument structure of acting vs. the one of appearing. A LM pre-trained using the SSP task can learn these implications, as it reasons about concepts from $s_3$, e.g., "appearing in Prison Break and Just Jordan" (which are TV series), being related to concepts from $s_2$, e.g., "having an acting career" as the sentences belong to the same paragraph. 

The semantics learned by connecting sentences in the same paragraph transfer well downstream, as the model can re-use previously learned relations between entities and concepts, and apply them between question and answer candidates. Relations in one sentence may be used to formulate questions that can be answered in the other sentence, which is most likely to happen for sentences in the same paragraph since every paragraph describes the same general topic from a different perspective. 

We design three ways of choosing the appropriate contextual information $c$ for SSP. We present details on how we sample spans $a$, $b$ and $c$ from the pre-training documents below.

\paragraph{Static Document-level Context (SDC)}
Here, we choose the context $c$ to be the first paragraph $P_0$ of $D=\{P_0,..,P_n\}$ from which $b$ is extracted. This is based on the intuition that the first paragraph acts as a summary of a document's content \cite{pretraining-large-retrieval-2020}: this strong context
can help the model at identifying if $b$ is extracted from the same paragraph as $a$. We call this static document-level context since the contextual information $c$ is constant for any $b$ extracted from the same document $D$. Specifically, the positive examples are created by sampling $a$ and $b$ from a single random paragraph $P_i \in D, i > 0$. For the previously chosen $a$, we create hard negatives by randomly sampling a sentence $b$ from different paragraphs $P_j \in D, j \ne i \wedge j > 0$. We set $c = P_0$ for this negative example as well since $b$ still belongs to $D$. We create easy negatives for a chosen $a$ by sampling $b$ from a random paragraph $P'_i$ in another document $D' \ne D$. In this case, $c$ is chosen as the first paragraph $P'_0$ of $D'$ since the context in the downstream AS2 task is associated with the answer candidate, and not with the question.

\paragraph{Dynamic Paragraph-level Context (DPC)}
We dynamically select the context $c$ to be the paragraph from which the sentence $b$ is extracted. We create positive examples by sampling $a$ and $b$ from a single random paragraph $P_i \in D$, and we set the context as the remaining sentences in $P_i$, i.e., $c = P_i \setminus \{a, b\}$. Note that leaving $a$ and $b$ in $P_i$ would make the task trivial. For the previously chosen $a$, we create hard negatives by sampling $b$ from another random paragraph $P_j \in D, j \ne i$, and setting $c = P_j \setminus \{ b \}$. We create easy negatives for a chosen $a$ by sampling $b$ from a random $P'_i$ in another document $D' \ne D$, and setting $c = P'_i \setminus \{ b\}$. 

\paragraph{Dynamic Sentence-level Local Context (DSLC)}
We choose $c$ to be the local context around the sentence $b$, i.e, the concatenation of the previous and next sentence around $b$ in $P \in D$. To deal with corner cases, we require at least one of the previous or next sentences of $b$ to exist (e.g., the next sentence may not exist if $b$ is the last sentence of the paragraph $P$). We term this DSLC as the contextual information $c$ is specified at sentence-level and changes correspondingly to every sentence $b$ extracted from $D$. We create positive pairs similar to SDC and DPC by sampling $a$ and $b$ from the same paragraph $P_i \in D$, with $c$ being the local context around $b$ in $P_i$ (and $a \notin c$). We automatically discard paragraphs that are not long enough to ensure the creation of a positive example. We generate hard negatives by sampling $b$ from another $P_j \in D, j \ne i$, while for easy negatives, we sample $b$ from a $P'_i \in D', D' \ne D$ (in both cases $c$ is set as the local context around $b$).

\begin{table*}[t]
\centering
\resizebox{\linewidth}{!}{
    
\begin{tabular}{lccccccccccccccc}
    \toprule
    \multirow{2}[3]{*}{\textbf{Model}} & \multirow{2}[3]{*}{\textbf{\shortstack{Context}}} & \multicolumn{2}{c}{\textbf{ASNQ}} & & \multicolumn{2}{c}{\textbf{WikiQA}} & & \multicolumn{2}{c}{\textbf{NewsAS2}} & & \multicolumn{2}{c}{\textbf{IQAD Bench 1}} & & \multicolumn{2}{c}{\textbf{IQAD Bench 2}} \\
    
    \cmidrule{3-4} \cmidrule{6-7} \cmidrule{9-10} \cmidrule{12-13} \cmidrule{15-16} &  & \textbf{MAP} & \textbf{P@1} & & \textbf{MAP} & \textbf{P@1} & & \textbf{MAP} & \textbf{P@1} & & \textbf{MAP} &\textbf{P@1} & & \textbf{MAP} &\textbf{P@1} \\

\midrule

ELECTRA-Base   & {\xmark}                               & 69.3 (0.0) & 65.0 (0.2)               & & 85.7 (0.9) & 78.5 (1.6)                 & & 81.3 (0.2) & 75.6 (0.2)                 & & \multicolumn{2}{c}{Baseline}        & & \multicolumn{2}{c}{Baseline} \\
ELECTRA-Base~$\clubsuit$ & {\cmark}                     & 72.3 (0.6) & 68.1 (0.8)               & & 83.1 (1.3) & 73.8 (2.1)                 & & 82.0 (0.4) & 76.0 (0.5)                 & & -0.6\% & -1.0\%                     & & -0.4\% & -0.9\% \\

\midrule

\textbf{(Ours)} ELECTRA-Base + SSP (SDC) & {\cmark}     & \bund{74.7} (0.5) & \und{69.6} (0.3)  & & \bund{88.7} (0.1) & \bund{82.9} (0.2)   & & \und{82.7} (0.2) & 77.0 (0.4)           & & \bund{+1.2\%} & \textbf{+0.6\%}     & & \bund{+0.9\%} & \bund{+1.4\%} \\
\textbf{(Ours)} ELECTRA-Base + SSP (DPC) & {\cmark}     & \und{74.4} (0.2)  & \bund{70.5} (0.2) & & \und{88.0} (0.6) & \und{81.3} (0.6)     & & 82.7 (0.5) & \bund{77.3} (0.7)          & & +0.4\% & -0.6\%                     & & +0.4\% & +0.1\% \\
\textbf{(Ours)} ELECTRA-Base + SSP (DSLC) & {\cmark}    & \und{74.3} (0.3)  & \und{70.0} (0.8)  & & \und{87.0} (0.9) & \und{79.7} (1.4)     & & \textbf{82.8} (0.4) & \bund{77.3} (0.5) & & \und{+1.0\%} & +0.6\%               & & \und{+0.2\%} & 0.0\%  \\
\textbf{(Ours)} ELECTRA-Base + SSP (All) & {\cmark}     & \und{73.8} (0.4)  & 68.8 (0.4)        & & \und{87.5} (0.5) & \und{81.5} (0.7)     & & \und{82.7} (0.2) & \und{77.2} (0.3)     & & +0.1\% & -0.4\%                     & & +0.1\% & -0.1\% \\

\midrule
\midrule

RoBERTa-Base & {\xmark}                                 & 68.2 (0.5) & 63.5 (0.5)               & & 85.1 (1.9) & 77.2 (3.1)                 & & 81.7 (0.1) & 76.2 (0.2)                 & & +0.6\% & +0.1\%                     & & +0.7\% & +1.3\% \\
RoBERTa-Base~$\clubsuit$  & {\cmark}                    & 71.6 (0.6) & 67.6 (0.6)               & & 84.4 (1.5) & 77.0 (2.1)                 & & 82.4 (0.2) & 76.6 (0.7)                 & & +0.4\% & 0.0\%                      & & +1.1\% & +1.7\% \\

\midrule

\textbf{(Ours)} RoBERTa-Base + SSP (SDC) & {\cmark}     & \und{73.1} (0.5) & 68.7 (0.8)         & & \und{87.8} (0.6) & \und{81.8} (0.9)     & & \und{82.8} (0.1) & 76.9 (0.2)           & & \bund{+1.7\%} & \bund{+3.0\%}       & & +1.0\% & +1.7\% \\
\textbf{(Ours)} RoBERTa-Base + SSP (DPC) & {\cmark}     & \bund{73.2} (0.4) & \bund{69.2} (0.5) & & \bund{89.9} (0.2) & \bund{85.2} (0.4)   & & 82.3 (0.1) & 76.0 (0.1)                 & & +0.4\% & +1.2\%                     & & +1.2\% & \bund{+2.7\%} \\
\textbf{(Ours)} RoBERTa-Base + SSP (DSLC) & {\cmark}    & \und{72.9} (0.4) & \und{69.0} (0.3)   & & \und{87.8} (0.9) & \und{81.6} (1.3)     & & 82.6 (0.2) & 77.0 (0.2)                 & & +0.6\% & +1.5\%                     & & +1.0\% & +1.4\% \\
\textbf{(Ours)} RoBERTa-Base + SSP (All) & {\cmark}     & 72.9 (0.6) & 68.2 (0.8)               & & \und{88.2} (0.9) & \und{82.4} (1.7)     & & \bund{83.0} (0.2) & \textbf{77.3} (0.5) & & \und{+1.2\%} & \und{+2.4\%}         & & \textbf{+1.4\%} & +2.2\% \\

    \bottomrule
\end{tabular}
}

\vspace{0.3em}
\caption{Results (std. dev. in parenthesis) on AS2. Models with $\clubsuit$ are from \cite{Lauriola2021}. {\cmark} and {\xmark} denote whether local contextual information was used in fine-tuning. SDC, DPC and DSLC indicate the pre-training variants of the SSP task that we propose.
Best results are in bold while we underline statistically significant improvements over the two contextual baselines ($\clubsuit$) using a Student \textit{t}-test with $95\%$ of confidence level.}
\vspace{0.3em}
\label{tab:results_as2}
\end{table*}

\section{Datasets}
\label{sec:datasets}
\vspace{-.2em}

\paragraph{Pre-Training}

To perform a fair comparison and avoid any improvement stemming from additional pre-training data, we use the same corpora as RoBERTa \cite{liu2019roberta}. This includes the English Wikipedia, the BookCorpus \cite{Zhu_2015_ICCV}, OpenWebText \cite{Gokaslan2019OpenWeb} and CC-News \footnote{STORIES is no longer publicly available, hence omitted}. We pre-process each dataset by filtering away: (i) sentences shorter than 20 characters, (ii) paragraphs shorter than 60 characters and (iii) documents shorter than 200 characters. We split paragraphs into sequences of sentences using the NLTK tokenizer~\citep{nltk} and create the SSP pre-training datasets following Section \ref{ssec:objectives}. Refer Appendix~\ref{app:pretraining_datasets} for more details.

\paragraph{Contextual AS2}
We evaluate our pre-trained models on three public and two industrial datasets for contextual AS2. 
For all datasets, we use the standard ``clean'' setting, by removing questions in the dev.~and test sets which have only positive or only negative answer candidates, following standard practice in AS2~\cite{garg2019tanda}. We measure performance using Precision-at-1 (P@1) and Mean Average Precision (MAP) metrics. 

\vspace{-0.75em}
\begin{itemize}[wide, labelwidth=0pt, labelindent=0pt]
    \itemsep-0.3em
    \item \textbf{ASNQ} is a large scale AS2 dataset \cite{garg2019tanda} derived from NQ~\cite{naturalquestion}. The questions are user queries from Google search, and answers are extracted from Wikipedia. 
    
    \item \textbf{WikiQA} is a small dataset~\cite{yang2015wikiqa} for AS2 with questions extracted from Bing search engine and answer candidates retrieved from the first paragraph of Wikipedia articles.

    \item \textbf{IQAD} is a large scale industrial dataset containing de-identified questions asked by users to Alexa virtual assistant. IQAD contains ${\sim}$220k questions where answers are retrieved from a large web index (${\sim}$1B web pages) using Elasticsearch. We use two different evaluation benchmarks for IQAD: (i) \emph{IQAD Bench 1}, which contains 2.2k questions with ${\sim}$15 answer candidates annotated for correctness by crowd workers and (ii) \emph{IQAD Bench 2}, which contains 2k questions with ${\sim}$15 answer candidates annotated with explicit fact verification guidelines for correctness by crowd workers. (Our manual analysis indicates a higher annotation quality for QA pairs in Bench 2 than Bench 1). Results on IQAD are presented relative to a baseline due to the data being internal.

    \item \textbf{NewsAS2} is a large AS2 dataset created from NewsQA \cite{trischler-etal-2017-newsqa}, a MR dataset, following the procedure of ~\citeauthor{garg2019tanda} for ASNQ. The dataset contains ${\sim}$70K human generated questions with answers extracted from \textit{CNN/Daily Mail}. More details about the procedure to create NewsQA are given in Appendix \ref{app:newsqa_datasets}.
\end{itemize}

\section{Experiments}
\label{sec:experiments}
\vspace{-.2em}

\paragraph{Continuous Pre-Training}
We use RoBERTa-Base and ELECTRA-Base
public checkpoints (pre-training from scratch would have required large amounts of computational resources), and perform continuous pre-training using our objectives for ${\sim}$10\% of the compute used by the original models. Complete details are given in Appendix \ref{app:continuous_pretraining}. We experiment with each of our pre-training objectives independently, as well as combining all of them.

\paragraph{Fine-Tuning}
We fine-tune each continuously pre-trained model on all the AS2 datasets. As baselines, we consider (i) standard pairwise-finetuned AS2 models, using only the question and the answer candidate, and (ii) contextual fine-tuned AS2 models from ~\cite{Lauriola2021}, which use the question, answer candidate and local context.

\section{Results}
\vspace{-.2em}

Table \ref{tab:results_as2} summarizes the results of our experiments averaged across 5 runs to show also standard deviation and statistically significant improvements over baselines.

\paragraph{Public datasets}

On ASNQ, our pre-trained models get 3.8 - 5.5\% improvement in P@1 over the baseline using only the question and answer. Our models also outperform the stronger contextual AS2 baselines (1.6\% with RoBERTa and 2.4\% with ELECTRA), indicating that our task-aware pre-training can help improve the downstream fine-tuning performance. On NewsAS2, we observe a similar trend, where all our models (except one) outperform both the standard and contextual baselines. On WikiQA, a smaller dataset, the contextual baselines under-performs the non-contextual baselines, highlighting that with few samples the model struggles to adapt and reason over three text spans. For this reason, our pre-training approaches provide the maximum accuracy improvement on WikiQA (up to 8 - 9.1\% over the non-contextual and contextual baselines).

\paragraph{Industrial datasets}

On IQAD, we observe that the contextual baseline performs on par or lower than the non-contextual baseline, indicating that off-the-shelf transformers cannot effectively exploit the context available for this dataset. The answer candidates and context for IQAD are extracted from millions of web documents. Thus, learning from the context in IQAD is a harder task than learning from it on ASNQ, where the context belongs to a single Wikipedia document. Our pre-trained models help to process the diverse and possibly noisy context of IQAD, and produce a significant improvement in P@1 over the contextual baseline.

\paragraph{Combining the 3 SSP objectives}

We observe that combining all the objectives together does not always outperform the individual objectives, which is probably due to the misalignment between the different approaches for sampling context in our pre-training strategies. Notice that we used a single classification head for all the three tasks, indirectly asking the model also to recognize the task to be solved among SDC, DPC or DSLC. Experiments with separate classification heads (one for each task) led to worse results in early experiments.

\paragraph{Choosing the optimal SSP objective}
 
Our fine-tuning datasets have significantly different structures: ASNQ, NewsAS2 and WikiQA have answer candidates sourced from a single document (Wikipedia for ASNQ and WikiQA and CNN Daily Mail articles for NewsQA), while IQAD has answer candidates extracted from multiple documents. This also results in the context for the former being more homogeneous (context for all candidates for a question is extracted from the same document), while for the latter the context is more heterogeneous (extracted from multiple documents for different answer candidates). 

Our DPC and DSLC pre-training approaches are well aligned in terms of the context that is used to help the SSP predictions. The former uses the remainder of the paragraph $P$ as context (after removing $a$ and $b$), while the latter uses the sentence previous and next to $b$ in $P$. We observe empirically that the contexts for DPC and DSLC often overlap partially, and are sometimes even identical (considering average length of paragraphs in the pre-training corpora is 4 sentences). This explains why models pre-trained using both these approaches perform comparably in Table~\ref{tab:results_as2} (with only a very small gap in P@1 performance).

On IQAD, we observe that the SDC approach of providing context for SSP outperforms DPC and DSLC. In SDC, the context $c$ can potentially be very different from $a$ and $b$ (as it corresponds to the first paragraph of the document), and this can aid exploiting information and effectively ranking answer candidates from multiple documents (possibly from different domains) like for IQAD. For these reasons, we recommend using DPC and DSLC when answer candidates are extracted from the same document, and SDC when candidates are extracted from multiple sources.

\section{Conclusion and Future Work}
\label{sec:conclusions}

In this paper, we have proposed three pre-training strategies for transformers, which (i) are aware of the downstream task of contextual AS2, and (ii) use the document and paragraph structure information to define effective objectives. Our experiments on three public and two industrial datasets using two transformer models show that our pre-training strategies can provide significant improvement over the contextual AS2 models.

In addition to local context around answer candidates (the previous and successive sentences), other contextual signals can also be incorporated to improve the relevance ranking of answer candidates. Meta-information like document title, abstract/first-paragraph, domain name, etc. corresponding to the document containing the answer candidates can help answer ranking. These signals differ from the previously mentioned local answer context as they provide ``global'' contextual information pertaining to the documents for AS2. Our SDC objective, which uses the first paragraph of the document for the context input slot, captures global information pertaining to the document, and we hypothesise that this may improve downstream performance using other global contextual signals in addition to local answer context.

\section*{Limitations}
Our proposed pre-training approaches require access to large GPU resources (pre-training is performed on 350M training samples for large language models containing 100's of millions of parameters). Even using 10\% of the original pre-training compute, the additional pre-training takes a long time duration to finish (several days even on 8 NVIDIA A100 GPUs). This highlights that this procedure cannot easily be re-done with newer data being made available in an online setting. However the benefit of our approach is that once the pre-training is complete, our released model checkpoints can be directly fine-tuned (even on smaller target datasets) for the downstream contextual AS2 task. For the experiments in this paper, we only consider datasets from the English language, however we conjecture that our techniques should work similarly for other languages with limited morphology. Finally, we believe that the three proposed objectives could be better combined in a multi-task training scenario where the model has to jointly predict the task and the label. At the moment, we only tried using different classification heads for this but the results were worse.

\bibliography{anthology,custom}
\bibliographystyle{acl_natbib}

\appendix
\clearpage

\section*{Appendix}

\section{Additional Dataset Details}

\subsection{Pre-Training Datasets}
\label{app:pretraining_datasets}

For each SSP objective, we randomly sample up to 2 hard negatives, and additionally sample easier negatives until the total number of negatives is 4. Instead of reasoning in terms of sentences, we design our SSP objectives to create $a$ and $b$ as small spans composed of 1 or more contiguous sentences. For $a$, we keep the length equal to 1 sentence because it emulates the question, which typically is just a single sentence. For $b$, we randomly assign a length between 1 and 3 sentences. The length of the context $c$ cannot be decided a-priori because it depends on the specific pre-training objective and the length of the paragraph. After the pre-processing, all the resulting continuous pre-training datasets contain around 350M training examples each.

\subsection{NewsQA dataset}
\label{app:newsqa_datasets}

We created NewsAS2 by splitting each document in NewsQA into individual sentences with the NLTK tokenizer~\citep{nltk}. Then, for each sentence, we assign a positive label if it contains at least one of the annotated answers for that document, and assign a negative label otherwise. The resulting dataset has 1.69\% positives sentences per query in the training set, 1.66\% in the dev set and 1.68\% in the test set.

\section{Frameworks \& Infrastructure}
\label{app:infrastructure}

Our framework is based on (i) HuggingFace Transformers \cite{wolf-etal-2020-transformers} for model architecture, (ii) HuggingFace Datasets \cite{lhoest-etal-2021-datasets} for data processing, (iii) PyTorch-Lightning for distributed training \citep{falcon2019pytorch} and (iv) TorchMetrics for AS2 evaluation metrics \cite{Detlefsen2022}. We performed our pre-training experiments for every model on 8 NVIDIA A100 GPUs with 40GB of memory each, using \ \textit{fp16} for tensor core acceleration.

\begin{table}[t]
\centering
\small
\resizebox{\linewidth}{!}{
    
\begin{tabular}{lccccccccc}
    \toprule
    \multirow{2}{*}{\textbf{Dataset}} & \multicolumn{2}{c}{\textbf{Train}} & & \multicolumn{2}{c}{\textbf{Dev}} & & \multicolumn{2}{c}{\textbf{Test}} \\ \cmidrule{2-3} \cmidrule{5-6} \cmidrule{8-9} & \textbf{\#Q} & \textbf{\#QA} & & \textbf{\#Q} & \textbf{\#QA} & & \textbf{\#Q} & \textbf{\#QA} \\

    \midrule
    ASNQ                    & 57242	& 20377568                                      && 1336 & 463914                                    && 1336 & 466148 \\
    \midrule
    WikiQA                    & 2118	& 20360                                      && 122 & 1126                                    && 237 & 2341 \\
    \midrule
    \multirow{2}{*}{IQAD}   & \multirow{2}{*}{221334} & \multirow{2}{*}{3894129}    && \multirow{2}{*}{2434} &
    \multirow{2}{*}{43369}   && 2252 & 38587 \\
                            & &                                                     && &                                                && 2088 & 33498 \\
    \midrule
    NewsAS2                  & 71561	& 1840533                                       && 2102 & 51844                                     && 2083 & 51472 \\
\bottomrule
\end{tabular}
}
\vspace{0.3em}
\caption{Number or unique questions and question-answer pairs in the fine-tuning datasets. IQAD Bench 1 and Bench 2 sizes are mentioned in the Test set column corresponding to IQAD.}
\label{tab:datasets_stats}
\end{table}

\section{Details of Continuous Pre-Training}
\label{app:continuous_pretraining}

We experiment with RoBERTa-Base and ELECTRA-Base public checkpoints. RoBERTa-Base contains 124M parameters while ELECTRA-Base contains 33M parameters in the generator and 108M in the discriminator.

We do continuous pre-training starting from the aforementioned models for 400K steps with a batch size of 4096 examples and a triangular learning rate with a peak value of $10^{-4}$ and 10K steps of warm-up. In order to save resources, we found it beneficial to reduce the maximum sequence length to 128 tokens. In this setting, our models see $\sim$210B additional tokens each, which is 10\% of what is used in the original RoBERTa pre-training. Our objectives are more efficient because the attention computational complexity grows quadratically with the sequence length, which in our case is 4 times smaller than the original RoBERTa model. 

We use cross-entropy as the loss function for all our pre-training and fine-tuning experiments. Specifically, for RoBERTa pre-training we add the MLM loss to our proposed binary classification losses using equal weights (1.0) for both the loss terms. For ELECTRA pre-training, we sum three loss terms: MLM loss with a weight of 1.0, the Token Detection loss with a weight of 50.0, and our proposed binary classification losses with a weight of 1.0.

During continuous pre-training, we feed the text tuples $(a,b,c)$ (as described in Section~\ref{ssec:objectives}) as input to the model in the following format: `$[\text{CLS}] a [\text{SEP}] b [\text{SEP}] c [\text{SEP}]$'. To provide independent sentence/segment ids to each of the inputs $a$, $b$ and $c$, we initialize the sentence embeddings layers of RoBERTa and ELECTRA from scratch, and extend them to an input size of 3.

The pre-training of every model obtained by combining ELECTRA and RoBERTa architectures with our contextual pre-training objectives took around 3.5 days each on the machine configuration described in Appendix \ref{app:infrastructure}. The dataset preparation required 10 hours over 64 CPU cores.

\begin{table}[t]
\centering
\small
\resizebox{\linewidth}{!}{
    
\begin{tabular}{llcccc}
    \toprule
    
    \textbf{Model} & \textbf{Hyper-parameter} & \textbf{ASNQ} & \textbf{WikiQA} & \textbf{NewsAS2} & \textbf{IQAD} \\

    \midrule

    \multirow{4}{*}{RoBERTa}    & Batch size        & 2048      & 32        & 256       & 256 \\
                                & Peak LR           & 1e-05     & 5e-06     & 5e-06     & 1e-05 \\
                                & Warmup steps      & 10K       & 1K        & 5K        & 5K \\
                                & Epochs            & 6         & 30        & 8         & 10 \\

    \midrule

    \multirow{4}{*}{ELECTRA}    & Batch size        & 1024      & 128       & 128       & 256 \\
                                & Peak LR           & 1e-05     & 2e-05     & 1e-05     & 2e-05 \\
                                & Warmup steps      & 10K       & 1K        & 5K        & 5K \\
                                & Epochs            & 6         & 30        & 8         & 10 \\

    \bottomrule
\end{tabular}
}
\vspace{0.3em}
\caption{Hyper-parameters used to fine-tune RoBERTa and ELECTRA on the AS2 datasets. The best hyper-parameters have been chosen based on the MAP results on the validation set.}
\label{tab:hparams}
\end{table}

\section{Details of Fine-Tuning}

The most common paradigm for AS2 fine-tuning is to consider publicly available pre-trained transformer checkpoints (pre-trained on large amounts of raw data) and fine-tune them on the AS2 datasets. Using our proposed pre-training objectives, we are proposing stronger model checkpoints which can improve over the standard public checkpoints, and can be used as the initialization for downstream fine-tuning for contextual AS2. 

To fine-tune our models on the downstream AS2 datasets, we found it is beneficial to use a very large batch size for ASNQ and a smaller one for IQAD, NewsAS2 and WikiQA. Moreover, for every experiment we used a triangular learning rate scheduler and we did early stopping on the development set if the MAP did not improve for 5 times in a row. We fixed the maximum sequence length to 256 tokens in every run, and we repeated each experiment 5 times with different initial random seeds. We did not use weight decay but we clipped gradients larger than 1.0 in absolute value. More specifically, for the learning rate we tried all values in $\{5*10^{-6}, 10^{-5}, 2*10^{-5}\}$ for RoBERTa and in $\{10^{-5}, 2*10^{-5}, 5*10^{-5}\}$ for ELECTRA. Regarding the batch size, we tried all values in $\{512, 1024, 2048, 4096\}$ for ASNQ, in $\{64, 128, 256, 512\}$ for IQAD and NewsAS2 and in $\{16, 32, 64, 128\}$ for WikiQA. More details about the final setting are given in Table \ref{tab:hparams}.

For the pair-wise models, we format inputs as `$[\text{CLS}] q [\text{SEP}] s_i [\text{SEP}]$', while for contextual models we build inputs of the form `$[\text{CLS}]q[\text{SEP}]s_i[\text{SEP}]c_i[\text{SEP}]$'. We do not use extended sentence/segment ids for the non-contextual baselines and retain the original model design: (i) disabled segment ids for RoBERTa and (ii) only using 2 different sentence/segment ids for ELECTRA. For the fine-tuning of our continuously pre-trained models as well as the contextual baseline, we use three different sentence ids corresponding to $q$, $s$ and $c$ for both RoBERTa and ELECTRA. Finally, differently from pre-training, in fine-tuning we always provide the previous and the next sentence as context for a given candidate.

The contextual fine-tuning of every models on ASNQ required 6 hours per run on the machine configuration described in Appendix \ref{app:infrastructure}. For other fine-tuning datasets, we used a single GPU for every experiment, and runs took less than 2 hours.

\section{Qualitative Examples}
\label{app:qualitative}

In Table \ref{tab:qualitative} we show a comparison of the ranking produced by our models and that by the contextual baselines on some questions selected from the ASNQ test set.

\makeatletter
\setlength{\@fptop}{0.5em}
\makeatother
\begin{table}[h]
\scriptsize
\begin{tabularx}{\linewidth}{cX}
    \toprule
    \multicolumn{2}{l}{\textbf{ELECTRA}} \\
    \midrule
    $\mathbf{Q}$ & \textbf{how many games does a team have to win for the world series} \\
    $\mathbf{A_1}$ & \textcolor{red}{Seven games were played, with the Astros victorious after game seven, played in Los Angeles.} \\
    $\mathbf{A_2}$ & In 1985, the format changed to best-of-seven. \\
    $\mathbf{A_3}$ & Since then, the 2011, 2014, and 2016 World Series have gone the full seven games. \\
    $\mathbf{A_4}$ & \textcolor{OliveGreen}{The winner of the World Series championship is determined through a best-of-seven playoff, and the winning team is awarded the Commissioner's Trophy.} \\
    $\mathbf{A_5}$ & The Houston Astros won the 2017 World Series in 7 games against the Los Angeles Dodgers on November 1st, 2017, winning their first World Series since their creation in 1962. \\
    \bottomrule
\end{tabularx}

\begin{tabularx}{\linewidth}{cX}
    \toprule
    \multicolumn{2}{l}{\textbf{RoBERTa}} \\
    \midrule

    $\mathbf{Q}$ & \textbf{where are trigger points located in the body} \\
    $\mathbf{A_1}$ & \textcolor{red}{Myofascial pain is associated with muscle tenderness that arises from trigger points, focal points of tenderness, a few millimeters in diameter, found at multiple sites in a muscle and the fascia of muscle tissue.} \\
    $\mathbf{A_2}$ & \textcolor{OliveGreen}{Myofascial trigger points, also known as trigger points, are described as hyperirritable spots in the fascia surrounding skeletal muscle.}\\
    $\mathbf{A_3}$ & \textcolor{ForestGreen}{Trigger points form only in muscles.} \\
    $\mathbf{A_4}$ & These in turn can pull on tendons and ligaments associated with the muscle and can cause pain deep within a joint where there are no muscles. \\
    $\mathbf{A_5}$ & They form as a local contraction in a small number of muscle fibers in a larger muscle or muscle bundle. \\
    \bottomrule
\end{tabularx}

\caption{Some qualitative examples from ASNQ test set where our ELECTRA and RoBERTa models with DSLC contextual continuous pre-training were able to rank the \textcolor{OliveGreen}{correct candidate} in the top position while the contextual baselines \textcolor{red}{failed}. The answer candidates are shown ranked by the ordering produced by the contextual baselines. Other positive candidates answers are colored in \textcolor{ForestGreen}{light green}.}
\label{tab:qualitative}
\end{table}

\end{document}